\documentclass{article}

\usepackage[nonatbib, final]{neurips_2019}

\usepackage[utf8]{inputenc} 
\usepackage[T1]{fontenc}    
\usepackage[colorlinks = true,
            linkcolor = blue,
            urlcolor  = blue,
            citecolor = blue,
            anchorcolor = blue]{hyperref}
\usepackage{url}            
\usepackage{booktabs}       
\usepackage{amsfonts}       
\usepackage{amsmath}
\usepackage{cases}
\usepackage{cite}
\usepackage{graphicx,subcaption,float}
\graphicspath{ {images/} }
\DeclareMathOperator{\sign}{sign}
\usepackage{nicefrac}       
\usepackage{microtype}      

\title{Distributed Soft Actor-Critic with Multivariate Reward Representation and Knowledge Distillation}

\author{%
  Dmitry Akimov\\
  \texttt{deakimov@edu.hse.ru} \\
}

\begin{document}

\maketitle

\begin{abstract}
    In this paper, we describe NeurIPS 2019 \textit{Learning to Move - Walk Around} challenge\footnote{\url{https://www.aicrowd.com/challenges/neurips-2019-learn-to-move-walk-around}} physics-based environment and present our solution to this competition which scored \textbf{1303.727} mean reward points and took \textbf{3rd} place.
    Our method combines recent advances from both continuous- and discrete-action space reinforcement learning, such as \textit{Soft Actor Critic} and \textit{Recurrent Experience Replay in Distributed Reinforcement Learning}. We trained our agent in two stages: to move somewhere at the first stage and to follow the target velocity field at the second stage. We also introduce novel Q-function split technique, which we believe facilitates the task of training an agent, allows critic pretraining and reusing it for solving harder problems, and mitigate reward shaping design efforts.
\end{abstract}

\section{Introduction}
    In this year participants were tasked to develop a controller for a musculoskeletal model with a goal to walk or run following velocity commands with minimum effort. Modern model-free deep reinforcement learning has demonstrated potential in many challenging domains, including past year competitions. The competition pushes forward researches from both reinforcement learning and neuromechanics fields.
    We believe that many clever ideas may arise from competitions like this. 

    This paper is organized as follows: at first we briefly describe the task and the environment with applicable reinforcement learning algorithms in section \ref{section:env_rl}, then we describe core algorithms in our solution in more details in section \ref{section:solution}, after that we present experiments and results in section \ref{section:experiments} and finally we discuss the results and conclude the work in section \ref{section:discussion}.

\section{Environment and Reinforcement Learning}\label{section:env_rl}
\begin{figure}
        \centering
        \includegraphics[width=1.0\textwidth]{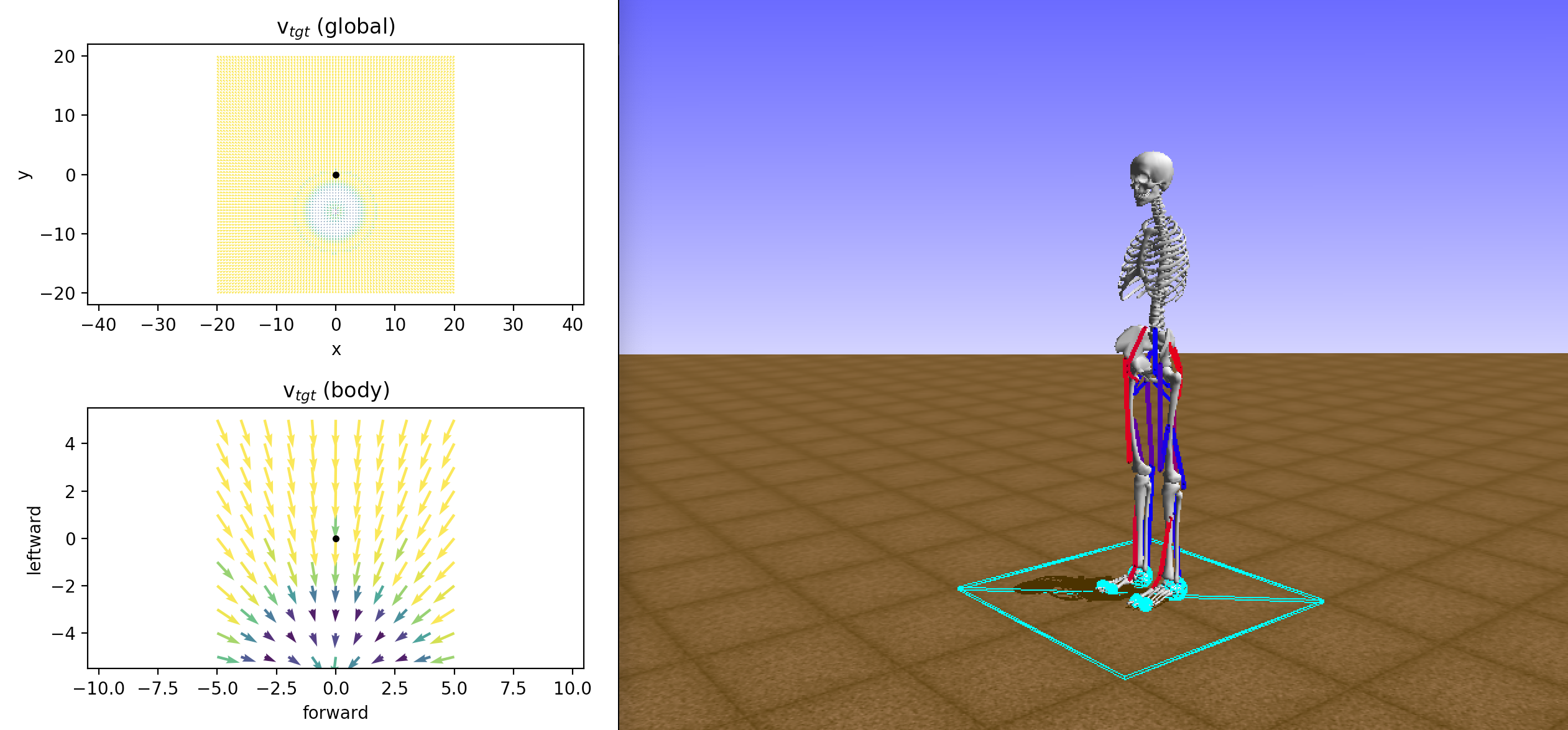}
        \caption{Environment screenshot. Global and local target velocity fields are plotted on the left side on the top and the bottom respectively. Local target velocity is relative to agent position which is marked as a black dot.}
        \label{fig:environment}
    \end{figure}

    The competition environment provides physiologically plausible 3D human model of a healthy adult, which consists of several segments for each leg, a pelvis segment, and a single segment to represent the upper half of the body (Figure \ref{fig:environment}). Environment provides vector $s \in \mathbb{R}^{97}$ which consists of pelvis state, ground reaction forces, joint angles and rates and muscle states as observation, and accepts vector $a \in \mathbb{R}^{22}$ as action, where each coordinate corresponds to a specific muscle. Also environment provides a local \textit{target velocity field}, randomly generated at the beginning of the episode - $2\times11\times11$ matrix, representing a $2D$ vector field on an $11\times11$ grid. The $2D$ vectors are target velocities, and the $11\times11$ grid is for every $0.5$ meters back-to-front and left-to-side.

    Environment provides reward function for the first competition round:
    \begin{equation}\label{equation:reward_fn}
        J(\pi) = R_{alive} + R_{step} 
               = \sum_i r_{alive} + \sum_{step_i} \left( 
                    w_{step} \cdot r_{step} - w_{vel} \cdot c_{vel} - w_{eff} \cdot c_{eff}
                \right)
    \end{equation}
    where \textit{alive} term forces agent to stay alive as long as possible and \textit{step} term forces agent to move towards the target direction with minimal effort. 
    There is an additional task bonus term in reward function for the second competition round, $R_{target}$, to ensure that agent 
    is able to move according to the target velocity field.
    More detailed environment description can be found on the
    environment page \footnote{\url{http://osim-rl.stanford.edu/docs/nips2019/environment/}}. Participants were expected to develop a controller for this human model via Reinforcement Learning (RL) \cite{sutton1998reinforcement}.

    The main task in Reinforcement Learning is to develop an agent or policy $\pi(a_t | s_t)$, which maximizes discounted sum of expected rewards: \begin{equation}\label{equation:rl_obj}
        J(\pi) = \sum_t \mathbb{E}_{(s_t, a_t) \sim \rho_\pi}[\gamma^t r(s_t, a_t)]
    \end{equation}
    where $s_t \in S$ is state, $a_t \in A$ is action, $r: S \times A \rightarrow [r_{min}, r_{max}]$ is reward function and by $\rho_\pi$ we denote the state-action marginals of the trajectory distribution induced by the policy $\pi(a_t|s_t)$. Two most popular and well-established algorithms for solving this task in continuous action spaces are \textit{Proximal Policy optimization} (PPO) \cite{schulman2017proximal} and \textit{Deep Deterministic Policy Gradient} (DDPG) \cite{lillicrap2015continuous} algorithms. While PPO provides very strong results in general, it is an \textit{on-policy} method, that requires new portion of data from environment at every optimization step and so its performance relies on the environment speed. On the other hand, DDPG is an \textit{off-policy} method, which means that it allows data re-usage for policy optimization. Current state-of-the-art DDPG improvement is \textit{Soft Actor-Critic} algorithm, \cite{haarnoja2018soft}\cite{haarnoja2018soft2}, which we briefly describe in section \ref{section:SAC} and which we used as the main RL procedure in the solution.

    Distributional training has shown its superiority to conventional approach in recent works \cite{barth2018distributed} \cite{horgan2018distributed} \cite{kapturowski2018recurrent}. All these works share key ideas about distributional approach, such as parallel data collection from many environment instances and prioritized experience replay. In our solution we adopted training pipeline from R2D2 \cite{kapturowski2018recurrent} and briefly describe it in section \ref{section:R2D2}.
\section{Solution}\label{section:solution}
    In this section we discuss existing techniques that we used as parts of the solution and explain in details novel multivariate reward representation technique.
    \subsection{Soft Actor-Critic} \label{section:SAC}
        Current state-of-the-art continuous control model-free reinforcement learning algorithm is \textit{Soft Actor-Critic} \cite{haarnoja2018soft} \cite{haarnoja2018soft2}. It is based on the maximum entropy framework, where agent aims to maximize expected reward while also maximizing entropy, or in other words, it aims to succeed at the task while acting as random as possible. This algorithm have proven its data efficiency and learning stability as well as hyper-parameter robustness, this is why we choose it as main RL algorithm in our solution. Formally, maximum entropy RL framework augments reward term in equation \ref{equation:rl_obj} with an entropy term:
        \begin{equation}\label{equation:sac_rl_obj}
            J(\pi) = \sum_t \mathbb{E}_{(s_t, a_t) \sim \rho_\pi}
            [
                \gamma^t 
                \left(
                    r(s_t, a_t) + 
                    \alpha \mathcal{H}(\pi(\cdot|s_t))
                \right)
            ]
        \end{equation}
        Where $\alpha$ determines the relative importance of the entropy term versus the reward, and thus controls the stochasticity of the optimal policy \cite{haarnoja2018soft2}. All known off-policy continuous action spaces methods in reinforcement learning relies on the actor-critic pair, where critic estimates Q-value:
        \[
            Q_\pi(s_t, a_t) = r(s_t, a_t) + \sum_{k=t+1}^{\infty}\mathbb{E}_{(s_k, a_k) \sim \rho_\pi}\left[\gamma^kr(s_k, a_k)\right]
        \] 
        function and actor proposes action, which maximizes Q-value function.

        In practice actor and critic are represented by neural networks $\pi_\phi(a_t|s_t)$ and $Q_\theta (s_t, a_t)$ with parameters $\phi$ and $\theta$. Standard practice is to estimate mean and variance of factorized Gaussian distribution, $\pi_\phi(a_t|s_t) = \mathcal{N}(\mu_\phi(s_t), \Sigma_\phi(s_t))$. Such choice of distribution allows to use reparametrization trick and train policy with via backpropagation. With such parametrization of actor and critic, learning objectives for actor, critic and entropy parameter writes as follows:
        \begin{equation}
            \begin{split}
                & J_\pi(\phi) = 
                    \mathbb{E}_{s_t \sim \mathcal{D}}
                    \left[
                        \mathbb{E}_{a_t \sim \pi_\phi}
                        \left[
                            \alpha  \log \left( \pi_\phi(a_t|s_t) \right) - Q_\theta(s_t, s_t)
                        \right]
                    \right], \\
                & J_Q(\theta) = \mathbb{E}_{(s_t, a_t) \sim \mathcal{D}}
                    \left[
                        \frac{1}{2} 
                        \left( 
                            Q_\theta(s_t, a_t) - 
                            (r(s_t, a_t) + \gamma \mathbb{E}_{s_{t+1}\sim p} \left[ V_{\bar{\theta}}(s_{t+1})\right]) 
                        \right)^2
                    \right], \\
                & J(\alpha) = \mathbb{E}_{a_t \sim \pi_t}[-\alpha \log \pi_t(a_t|s_t) - \alpha \bar{\mathcal{H}}],
            \end{split}
        \end{equation}
        where experience replay is denoted by $\mathcal{D}$ and objectives may be optimized by any stochastic gradient descent method. In addition to ones already mentioned, policy has another decent property: it constantly explores promising actions while giving up on clearly bad ones. 
    \subsection{Recurrent Experience Replay in Distributed Reinforcement Learning} \label{section:R2D2}
        One of the main differences between supervised and reinforcement learning is that training data must be fully processed during training, which means that reinforcement learning pipeline includes data collecting, data storing, data sampling to train an agent. Recent advances suggests to separate all this stages from each other and collect data by many parallel actors running on separate environment instances. We adopt this techniques from R2D2 and briefly describe them below.
        \subsubsection{Data collection}
\begin{figure}
                \centering
                \includegraphics[width=0.8\textwidth]{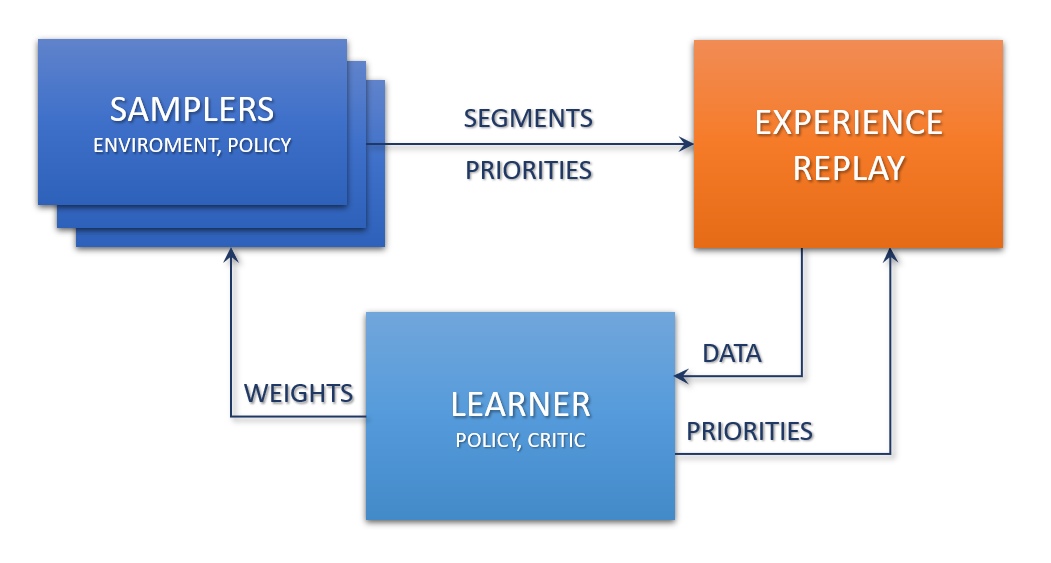}
                \caption{R2D2 training pipeline. Multiple samplers collect data from its own environment copy in parallel and send it into experience replay with initial priorities. The single learner samples data from experience replay to train on and periodically updates policy weights inside samplers.}
                \label{fig:r2d2_pipeline}
            \end{figure}

            Taking actions and collecting experience process is separated from learning in distributed RL context. In practice there are many actors (workers) collecting experience and one learner which trains the agent. The number of workers usually is high enough to generate data faster than learner can process it all, so there is another algorithm to select only viable data to train on.
            Full learning pipeline is shown at figure \ref{fig:r2d2_pipeline}
        \subsubsection{Data storage}
            Traditional data storage for an off-policy reinforcement learning algorithm is \textit{Experience Replay} (ER). It is a storage where agent sends transactions $(s, a, r, s')$ during collecting data from environment and samples data to learn from during training. \textit{Prioritized Experience Replay} \cite{schaul2015prioritized} changes the probability of sampling transactions according to corresponding critic loss on these transaction, so transactions with higher loss value are more likely to be sampled for training. R2D2 approach suggests to store in ER not the transaction itself, but overlapping sequences of consecutive $(s, a, r)$. Adjacent sequences overlap each other by half time steps and never cross episode boundaries. We refer to these sequences as segments.

            To compute prioritization weights over segments, R2D2 pipeline uses $n$-step prioritization based on $n$-step TD-errors $\delta_i$ over the sequence: $p = \eta \; \max_i \delta_i + \left( 1 - \eta \right) \bar{\delta}$, where $\eta$ is set to $0.9$.
        \subsubsection{Optimization}
            For loss computation and optimization purposes an invertible value function rescaling of the form
            $h(x) = \sign(x)\left(\sqrt{|x| + 1} - 1 \right) + \epsilon x$
            is used, which resulted in the following $n$-step targets for the Q-value function:

            \begin{equation}
                \begin{split}
                    & \hat{y_t} = h
                    \left(
                        r_{t+n-1} + \gamma ^ n h^{-1}(Q(s_{t+n}, a^*))
                    \right)\\
                    & r_{t+n-1} = \sum_{k=0}^{n-1} \gamma^k r_{t+k}
                \end{split}
                \notag
            \end{equation}{}

            The main contribution of R2D2 paper is the investigation of training agents with recurrent model in distributional framework. However, we did not use recurrent models in our agent, so we do not describe details unrelated to our solution.
    \subsection{Multivariate Reward Representation}
        Designing a reward function that leads to desired behaviour is known to be a challenging task in reinforcement learning. Usually environment represents high-dimensional non-linear dynamic system by itself, so even small changes in reward function may change optimal policy substantially and lead to unexpected behaviour. It has become a standard approach to use reward shaping (i.e. adapt reward signal). For example, if one wants agent to move forward with minimal effort, they provide some velocity bonus and sum it with effort penalty, so different reward parts will interfere and influence each other.

        Another problem is that it is hard or impossible to add additional reward term or modify existing ones. For example, if after several initial experiments resulted agents learned how to make steps but didn't learn how to move forward, the only way is to add new bonus-for-moving term into reward and hope that agent eventually learn or retrain agent from scratch. Both approaches lead to information forgetting, which is undesirable, especially in slow- or cost-sampling environments.

        To tackle with this problems, we introduce novel Q-function split technique which we call Multivariate Reward Representation. Suppose that scalar reward function may be decomposed into weighted sum of $n$ terms as 
        \begin{equation}\label{equation:mvrr}
            r_t = \sum_{i=1}^n w_i \cdot r_{i,t}
        \end{equation}
        at each time step $t$. Classic approach in reinforcement learning is to just sum reward and optimize corresponding Q-function. However, it may be beneficial to use each term separately and fit personal Q-function for each. First of all, with this representation reward terms do not interfere with each other. Secondly, it allows critic pretraining - the ability to add new or remove any existed reward terms. In practice critic is represented by deep neural network, in which the last layer is parametrized by matrix $\theta_{last} \in \mathbb{R}^{n \times m}$ (where hidden dim denoted by $m$) and projects features $x \in \mathbb{R}^m$ onto Q-values: $\bar{Q} = \theta_{last} \cdot x = {[Q_1, \ldots, Q_n]}^T \in \mathbb{R}^n$. To add a new reward term with such parametrization one may extend parameter matrix $\theta_{last}$ by one row. Removing reward term may be done by excluding corresponding row from parameter matrix or by setting corresponding weight to zero. And finally, it mitigate reward shaping design effort, because it is easier for critic to distinguish different reward terms from each other (it may be thought as of adding auxiliary tasks), but weights $w_i$ should be properly tuned anyway.

        Training both actor and critic with multivariate reward representation is straightforward. Vector form (with one coordinate per reward term) of critic loss should be optimized. Actor may optimize policy as usual with Q-function which is represented as scalar: 
        \[
            Q(s_t, a_t) = \sum_{i=1}^n w_i \cdot Q_i (s_t, a_t)
        \]
        also it is possible to optimize several actors with a set of different weights $\{w_i\}$ at once, but we'll leave investigation for further works.

        In our solution we used several different reward functions, which we will describe in details in the next section. 

    \subsection{Reward shaping}
        Our vectorized reward function consists of 7 different terms and has the following form:
        \begin{equation}\label{equation:reward_vector}
            \bar{r} = 
            \left[
                r_{env}, r_{clp}, r_{vdp}, r_{pvb}, r_{dep}, r_{tab}, r_{entropy}
            \right]
        \end{equation}
        At the early training stages agent tended to cross its legs, so we constructed \textit{crossing legs penalty}:
        \[
            r_{clp} = \min(0, (\mathbf{r}^{head} - \mathbf{r}^{pelvis}, \mathbf{r}^{left} - \mathbf{r}^{pelvis}, \mathbf{r}^{right} - \mathbf{r}^{pelvis}))
        \] where $\mathbf{r}$ is a radius vector and the second term inside minimum operator is scalar triple product of three vectors.
        To force our agent to follow target velocity, we applied \textit{velocity deviation penalty}
        \[
            r_{vdp} = -\sum_{i \; in \; step_i} \|v_{body} - v_{tgt}\|.
        \]
        At the early training our agents refused to move, so we added \textit{pelvis velocity bonus} 
        \[
            r_{pvb} = \|v_{body}\|
        \] 
        to force them to move.
    
        To force agent to move with minimal effort we added \textit{dense effort penalty}:
        \[
            r_{dep} = -\|action_t\|.
        \]
        To force agent to stand in the center of the target velocity field we added \textit{target achieve bonus}: 
        \begin{numcases}{r_{tab}=}
            0,               & $0.7 < \|v_{tgt}\|$ \notag \\
            0.1,             & $0.5 < \|v_{tgt}\| \leq 0.7$ \notag \\
            1 - 3.5 \|v_{tgt}\|^2, & $      \|v_{tgt}\| \leq 0.5$ \notag
        \end{numcases}
    
        And the last reward coordinate is entropy bonus from SAC: $r_{entropy} = \alpha * \mathcal{H}(\pi(\cdot|s_t))$. In addition we tried to add \textit{bending knees bonus}, however agents either quickly got stuck with the constantly bending policy if corresponding weight was too high, or just ignored this bonus if corresponding weight was too low, so we removed this bonus from training.
    \subsection{Full Agent}
        Our full agent combines all mentioned methods, such as loss functions and networks from SAC, learning pipeline i.e. parallel data collection (but on single-machine, in multi-threaded context) and prioritization, as well as $n$-step Q-learning and invertible value function rescaling from R2D2 and proposed Multivariate Reward Representation.

\section{Experiments}\label{section:experiments}
    In this section we describe architecture details and hyperparameters, as well as conducted experiments that are part of our solution.
    \subsection{Walk-forward agent pretraining}\label{section:pretraining}
        Original environment goal - walk and follow target velocity with minimum effort - is too hard for randomly initialized agent. So we decided to divide this task in two: to learn to move in \textit{any} direction with high speed for the beginning and only after that to learn to move in required direction.

        To achieve the first goal, we used 4-layer perceptron for both policy and critic networks, with input size of $\dim(S) = 97$ for policy and $\dim(S) + \dim(A) = 97 + 22 = 119$ for critic, hidden size of 256, layer norm before activation function, 'ELU' activation for policy and 'ReLU' for critic, and residual connections. We used 2 critics as proposed in \cite{fujimoto2018addressing} and took minimum value from them for Q-function estimation. We did not explore large variety of training hyperparameters, such as gamma value, learning rate, number and size of layers etc. due to the time limits and environment slowness. We have chosen discount factor $\gamma$ to be equal to $0.99$ at the beginning of our experiments.  Experience replay size was set to $250.000$ with segment length $10$. We set number of parallel data samplers and environment instances to 30. Adam optimizer \cite{kingma2014adam} was used with learning rate $3\cdot10^{-5}$ for policy and $1\cdot10^{-4}$ for critic. Batch size was equal to 256 and segment length to 10. Priority exponents $\alpha$ and $\beta$ were set to $0.1$ at the beginning of training and linearly increased to $0.9$ in $3000$ training steps. Our learning agent performed roughly 16 training steps per 1 new segment collected by one worker. Both policy and critic networks didn't get target velocity as input.

        We set reward weights vector in equation \ref{equation:mvrr} to 
        $\bar{w} = [1, 10, 0, 1, 1, 0, 1]$, and set $w_{vel} = 0$ in the environment reward $r_{env}$, equation \ref{equation:reward_fn}.
        With these weights it is beneficial to agent to move forward with maximal velocity and minimal effort in any direction. SAC encourages random behavior, so policy trained with these weights will move in random direction. With DDPG we cannot achieve such policy, because resulted agent will choose actions deterministically and follow only one trajectory.

        \begin{figure}
            \centering
            \includegraphics[width=0.65\textwidth]{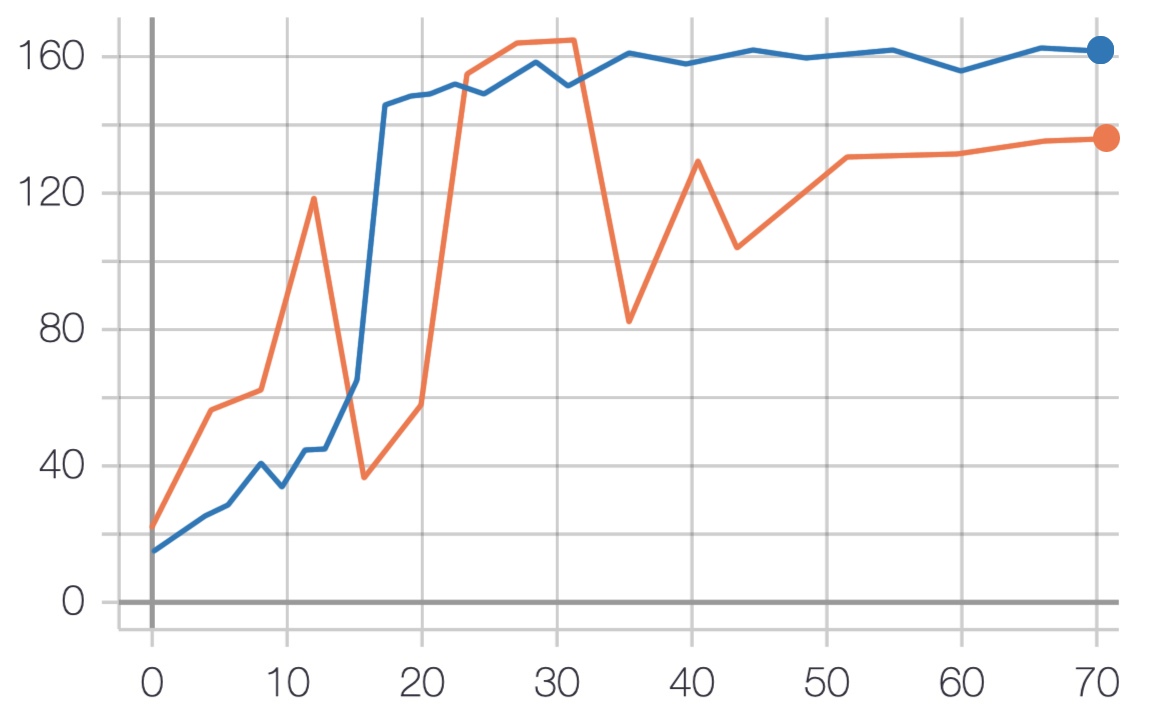}
            \caption{Test environment reward during pretraining for agents with Multivariate Reward Representation (blue) and without (orange). The horizontal axis represents time (in hours) and the vertical axis represents environment reward with $w_{vel} = 0$ obtained on difficulty 0 and averaged over 5 episodes.}
            \label{fig:pretrain_reward}
        \end{figure}

        Pretraining result may be seen at figure \ref{fig:pretrain_reward}. We test agents on 5 episodes once per training epoch, for testing we choose action deterministically as $a_t = \mu_\phi(s_t)$ and plot average test reward. 

        To prove that the proposed multivariate reward representation is beneficial, we decided to test agent in exactly the same training setup and reward function, but with the conventional one-headed critic and summed scalar reward. We tracked agents behaviour during training and observed that the agent without vectorized reward firstly learned to just stand with minimal effort (first 30 hours) and to make small ground-touching steps by one (right) of his leg. This agent started to learn how to move only after several additional training epochs, which can be seen at the reward-drop point after 30 hours. In contrast, the agent with vectorized reward did not suffer from getting stuck in sub-optimal policy and started to learn to move from the beginning of training. Plots caped at the point after which testing rewards did not increased substantially. Policy trained without vectorized reward converged to approximately 140 reward and policy with vectorized reward converged to approximately 160 reward. It is also worth to mention that the agent with Multivariate Reward Representations learned much faster and stable than the other one.
    \subsection{Knowledge distillation}
        After our agent successfully learned how to walk in random direction, we decided to add target velocity sensors into policy and critic in order to learn the initial task. Transferring knowledge from pretrained actor and critic (teachers) to the new ones - target velocity aware (students) - was straightforward. We trained newly initialized models $\pi_\phi^{s}(a_t|s_t,v_t)$ and $Q_\theta^{s}(s_t, v_t, a_t)$ by minimizing Kullback-Leibler divergence between policies and mean squared error between critics on data from previously saved experience replay:
        \begin{equation}\label{equation:distillation_loss}
            \begin{split}
                & J_{\pi^s}(\phi) = 
                    \mathbb{E}_{s_t \sim \mathcal{D}}
                    \mathbb{E}_{v_t \sim \mathcal{N}(0, 0.1)}
                    \left[
                        D_{KL}\left(\pi_\phi^{s}(a_t|s_t, v_t)||\pi_{\phi'}^t(a_t|s_t)\right)
                    \right] \\
                & J_{Q^s}(\theta) = 
                    \mathbb{E}_{s_t \sim \mathcal{D}}
                    \mathbb{E}_{v_t \sim \mathcal{N}(0, 0.1)}
                    \left(
                        Q_\theta^s\left(s_t, v_t, a \sim \pi^{s}\left(.|s_t, v_t\right)\right) - 
                        Q_{\theta'}^t(s_t, a \sim \pi^{t}(.|s_t))
                    \right)^2
            \end{split}
        \end{equation}
        Because we discarded target velocity during pretraining, we decided to sample random noise with small amplitude as a target velocity during distillation. As a result, distilled agent was aware of target velocity but ignored it and moved in random direction, just as teacher.

        Models $\pi^s(a_t|s_t,v_t)$ and $Q^s(s_t, v_t, a_t)$ share the same architecture with $\pi^t(a_t|s_t)$ and $Q^t(s_t, a_t)$ except input dim is now $\dim(S) + \dim(V) = 97 + 2\cdot11\cdot11=339$ for policy and $\dim(S) + \dim(V) + \dim(A) = 339 + 22 = 361$ for critic and hidden size equals to 1024 for both. We optimize distillation losses (equations \ref{equation:distillation_loss}) with Adam optimizer \cite{kingma2014adam} with learning rate $1\cdot10^{-4}$ for policy and critic networks and batch size 128.

        \begin{figure}
            \centering
            \includegraphics[width=0.6\textwidth]{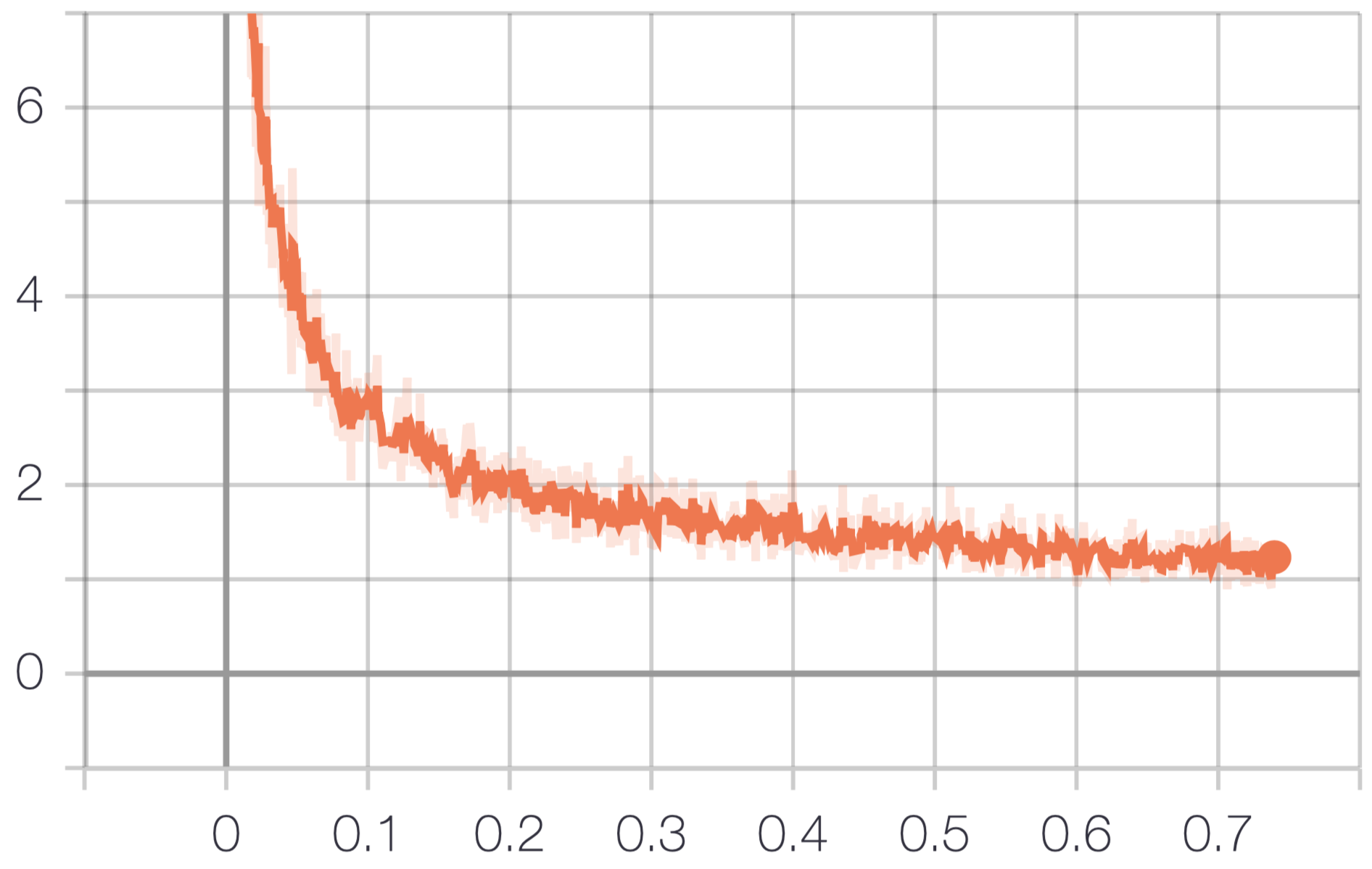}
            \caption{Policy distillation loss. The horizontal axis represents time (in hours) and the vertical axis represents $D_{KL}$ value.}
            \label{fig:kl_loss}
        \end{figure}

        Policy loss for described knowledge distillation process may be seen on figure \ref{fig:kl_loss}. The distillation process took only 44 minutes and resulted agent produced almost identical policy in terms of $D_{KL}$, and visually teacher and student behaviours were indistinguishable.

    \subsection{Main agent training}
        The agent that we have had so far could generate various trajectories in the target velocity field. So agent will experience positive and  negative reward and will explore effectively. We switched reward weights in equation \ref{equation:mvrr} to $\bar{w} = [1, 10, 1, 1, 1, 1, 1]$, and also changed $r_{pvb} = \cos(\theta)\| v_{body}\|$ where $\theta$ is angle between $v_{tgt}$ and $v_{body}$ vectors. With these changes agent received smaller velocity bonus if it moves in wrong direction, penalty for deviating from target velocity and huge bonus for target achieving. We start learning with new empty experience replay, and believe that with experience prioritization it quickly learned only in viable to critic data. This whole process resembles \textit{Hindsight Experience Replay}\cite{andrychowicz2017hindsight} algorithm, but much simpler and agent (or, more correctly, loss prioritization) chooses good data for training and does this automatically. All training parameters were the same as in the \ref{section:pretraining} section.

        We have tuned our agent two times: on difficulty 2 for round 1 and on difficulty 3 for round 2. 
        Agent converges to approximately 165 reward on difficulty 2 in 46 hours.

\begin{figure}
            \centering
            \includegraphics[width=0.8\textwidth]{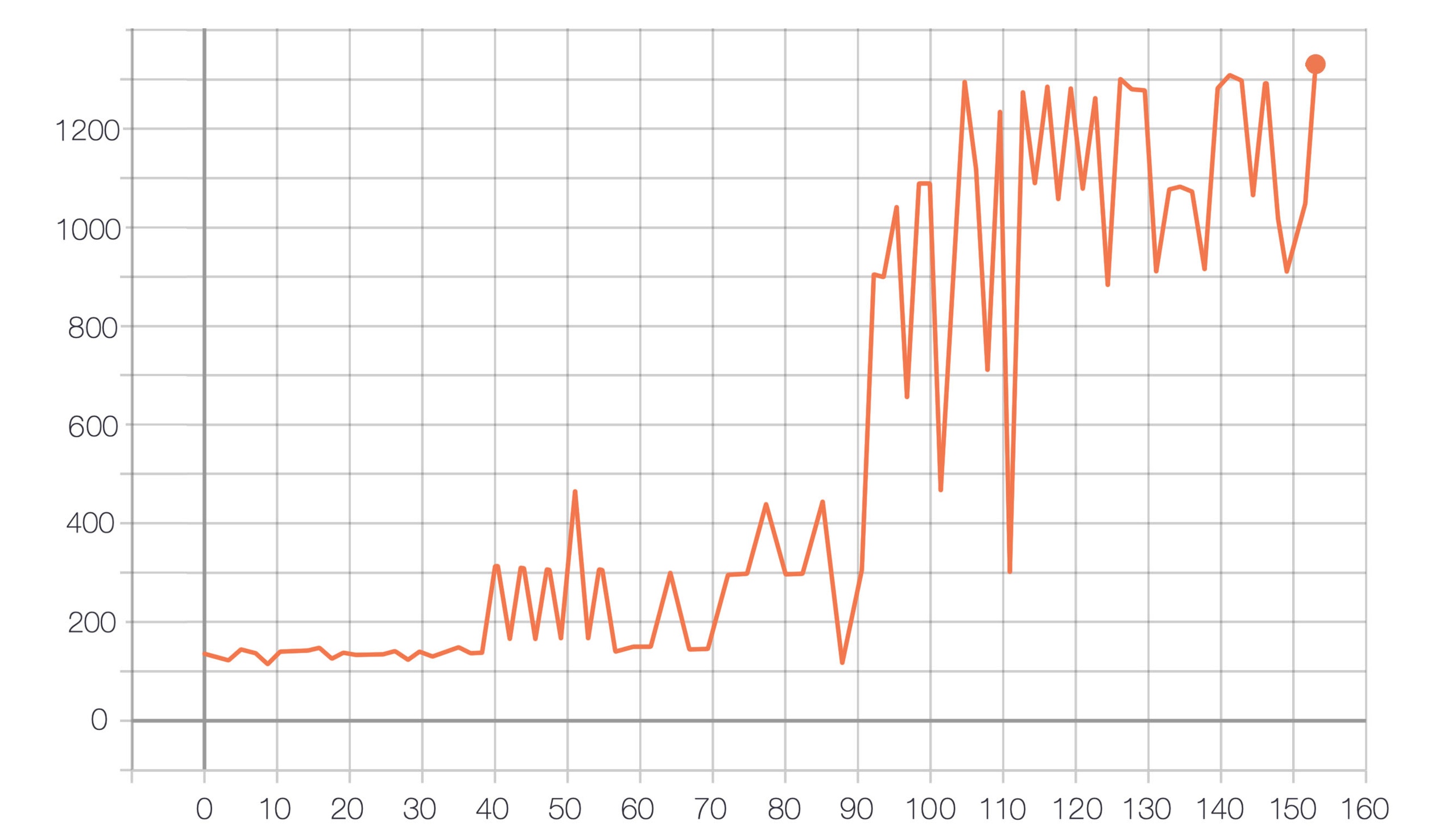}
            \caption{Test environment reward during tuning on difficulty 3.}
            \label{fig:tune_3}
        \end{figure}

        However, main difficulties arise at round 2 and difficulty 3, where agent must stand in the center of the target velocity. Our agent only knew how to approach the center but never experienced reward for standing inside. As may be seen on figure \ref{fig:tune_3}, training on difficulty 3 took 160 additional hours before
        the agent learned well how to stand in the center of target velocity.

\section{Discussion}\label{section:discussion}
    We proposed sample-efficient solution to the NeurIPS2019 Learn to Move competition. It combines several reinforcement learning techniques that, we believe, successfully complement each other, but careful ablation study is required to determine the importance of each component. Our final single-model policy is capable to move towards the target velocity field and get both bonuses. 

    It is worth mentioning that full pretraining and training processes from section \ref{section:experiments} may be done by \textit{Latent Space Policy for Hierarchical RL} \cite{haarnoja2018latent} in more end-to-end manner as well, however our approach with knowledge distillation is simple, fast and produced sufficient policy.
    Our code and policy video will be released soon and will be available on github: \url{https://github.com/DEAkimov/learn_to_move}.

    Key features of the solution: feed forward agent, Soft Actor Critic, parallel actors, knowledge distillation, Multivariate Reward Representation.
\section*{Acknowledgements}
    We thank our colleagues Ivanov Daniil, Markovich Alexander, Petrova Anastasia for the inspiration and motivation during the competition, additional thanks to Chesnokov German and Khodyreva Viktoriia for the text review.
    Thanks in particular to Maxim Galayko for helpful conversations and advice during development and for text review.

    Thanks to \textit{Mail.ru Group} company for providing computing resources.

\bibliographystyle{unsrt}
\bibliography{sample}
\end{document}